\documentclass[letterpaper, 10 pt, conference]{ieeeconf}  

\IEEEoverridecommandlockouts                              

\usepackage{graphicx}
\usepackage{float}
\usepackage{subfig}

\usepackage{verbatim}
\usepackage{xfrac}
\usepackage{mathtools}
\usepackage{times}
\usepackage{epsfig}
\usepackage{amsmath}
\usepackage{amssymb}
\usepackage{stfloats}
\usepackage{combelow}
\usepackage{flushend}
\usepackage{gensymb}
\usepackage{booktabs, multicol, multirow}
\usepackage{comment}
\usepackage{combelow}
\usepackage[pagebackref=true,breaklinks=true,letterpaper=true,colorlinks,bookmarks=false]{hyperref}

\DeclareMathOperator*{\argmin}{argmin} 

\title{\LARGE \bf Monocular Depth Estimation for Soft Visuotactile Sensors}

\author{
Rare\cb{s} Ambru\cb{s}* \quad 
Vitor Guizilini* \quad
Naveen Kuppuswamy* \\
Andrew Beaulieu \quad 
Adrien Gaidon \quad 
Alex Alspach \\
\vspace{-2mm} \\
Toyota Research Institute (TRI) \\
\texttt{first.lastname@tri.global} \\
\thanks{*Authors contributed equally.}
}

\begin{document}
\maketitle
\thispagestyle{empty}
\pagestyle{empty}


\begin{abstract}
Fluid-filled soft visuotactile sensors such as the \emph{Soft-bubbles} alleviate key challenges for robust manipulation, as they enable reliable grasps along with the ability to obtain high-resolution sensory feedback on contact geometry and forces. Although they are simple in construction, their utility has been limited due to size constraints introduced by enclosed custom IR/depth imaging sensors to directly measure surface deformations. Towards mitigating this limitation, we investigate the application of state-of-the-art monocular depth estimation to infer dense internal (tactile) depth maps directly from the internal single small IR imaging sensor. Through real-world experiments, we show that deep networks typically used for long-range depth estimation (1-100m) can be effectively trained for precise predictions at a much shorter range (1-100mm) inside a mostly textureless deformable fluid-filled sensor. We propose a simple supervised learning process to train an object-agnostic network requiring less than 10 random poses in contact for less than 10 seconds for a small set of diverse objects (mug, wine glass, box, and fingers in our experiments). We show that our approach is sample-efficient, accurate, and generalizes across different objects and sensor configurations unseen at training time. Finally, we discuss the implications of our approach for the design of soft visuotactile sensors and grippers\footnote{Code:~\href{https://github.com/TRI-ML/packnet-sfm}{https://github.com/TRI-ML/packnet-sfm}}.


\end{abstract}


\section{Introduction}
\label{sec:intro}

Visuotactile sensors or camera-based tactile sensors are a somewhat recent innovation in manipulation and are based on the principle that the effects of a tactile interaction between a robot and a surface can be observed by embedding a camera within a compliant surface \cite{Shimonomura2019}. Their ability to capture high-resolution contact geometry and interaction forces when a robot interacts with the world presents huge advantages in manipulating in cluttered and complex environments such as homes.   While several variants of such sensors have been reported for various kinds of robot manipulation applications (cf.~\cite{Yamaguchi2019} for a recent review), there are inherent trade-offs in mechanical complexity, richness of the sensed tactile information, the ease of calibration, and the complexity of resultant data processing. A key sensory requirement on these types of sensors is the need to capture high-resolution tactile depth maps that can then be utilized in various feedback control strategies for robust manipulation.

\begin{figure}[t!]
\centering
\subfloat[KITTI \cite{geiger2013vision}]{
\includegraphics[width=0.14\textwidth,height=4.0cm]{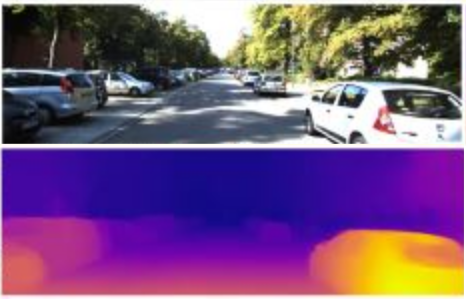}
}
\subfloat[NYUv2 \cite{nyuv2}]{
\includegraphics[width=0.14\textwidth,,height=4.0cm]{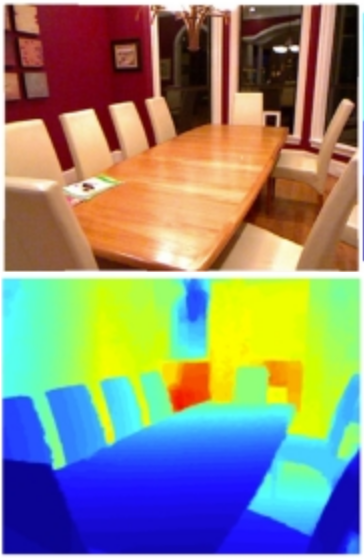}
}
\subfloat[Visuotactile (ours)]{
\includegraphics[width=0.14\textwidth,,height=4.0cm,trim=0 0 250 0,clip]{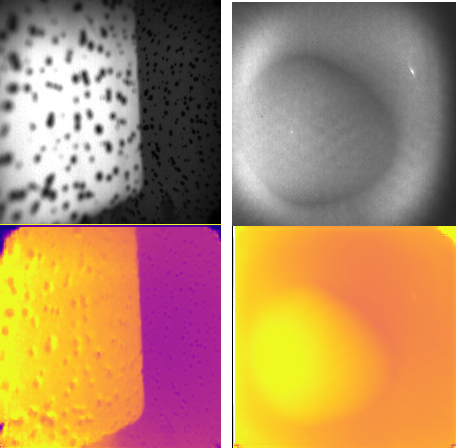}
}

\caption{Examples of (a) outdoor and (b) indoor datasets on which monocular depth estimation networks are usually used, and (c) our proposed visuotactile dataset.}
\vspace{-5mm}
\label{fig:teaser}
\end{figure}

From the standpoint of grasp stability -- a key requirement in several home manipulation tasks -- it is well known that equipping grippers and end-effectors with the appropriate degree of compliance is vital. While several examples of soft grippers have been reported in literature \cite{Hughes2016}, a recently proposed archetype, the \emph{Soft-bubble} gripper \cite{kuppuswamy_soft-bubble_2020} presents an interesting approach that combines highly compliant gripping surfaces on the fingers with visuo-tactile sensing by embedding an imaging sensor within an elastic membrane surface \cite{Alspach2019}. This approach results in perceptive grippers that can not only capture multi-modal tactile information on the contacts but in a mechanically robust and simple design that can be inexpensively constructed. An important modality captured by these sensors is that of tactile depth maps which are key towards capturing contact geometry - the large range of deformation exhibited by these sensors make this a challenging design problem that has been solved so far with COTS imaging sensors that have range and field-of-view limitations. Finding computationally efficient methods to obtain this kind of information is key in mitigating the limitations posed by the internal imaging system and thus expanding the utility of this sensor and gripper. 

Depth maps resulting from contact have been obtained from tactile sensors using three broad strategies: (i) model based methods using photometric stereo (e.g., \emph{gelsight} \cite{yuan2015measurement}), (ii) direct methods by using internal depth sensors (e.g., \emph{soft-bubbles} \cite{Alspach2019}), and (iii) inference on the RGB images using machine learning (e.g., \emph{gelslim} \cite{bauza2019tactile}). The choice of the depth estimation method is closely tied to the mechanical design of the visuotactile sensor itself, which in turn can affect other design factors such as sensor size, complexity of external data processing, and mechanical robustness. Furthermore, in the case of depth inference using machine learning, techniques demonstrated thus far (e.g., in \cite{bauza2019tactile} and \cite{yuan_gelsight_2017}) usually raise a requirement for a dedicated calibration/ground-truth generation procedure that may require synthesis of precise tactile imprints. Finding more efficient ways of acquiring this tactile depth map is key to enabling transferability and ease of use of such sensors, which is a primary motivation of this work. In the case of \emph{Soft-bubbles}, a key design limitation is the utilization of an internal Time-of-Flight (ToF) depth sensor -- the complexities of depth measurement in such close ranges (difficulty in meeting the required optical specifications) limit the miniaturization of the overall form factor of \emph{Soft-bubbles}. Replacing this imaging system with a simpler one improves this overall design and enables miniaturization, as long as reasonable tactile depth can be inferred from the resulting images, as we demonstrate in this paper. 



We therefore tackle this problem through our \textbf{primary contribution} of adapting a  state-of-art method in deep learning for monocular depth estimation to estimate per-pixel depth. From the standpoint of the monocular depth estimation problem, the sensing range requirements in the \emph{Soft-bubbles} represents a completely different domain than usually explored, both visually (cf.~Figure~\ref{fig:teaser}) and range-wise by three orders of magnitude (millimeter-level vs meter-level precision). Our approach uses the IR/depth image pairs to train a deep convolutional network that regresses per-pixel depth. 

Our \textbf{second contribution} is a sample efficient training process requiring only a small candidate set of contacted object geometries coming in contact for just a few minutes in total. We show this procedure is sufficient to learn a depth network that generalizes to novel poses and object categories. Furthermore, our trained models generalize well across multiple sensors (i.e., across multiple internal cameras and different pseudorandom textured membranes).

Our \textbf{third contribution} is to show that the learned models exhibit similar-or-better noise characteristics when compared to the existing "ground truth" depth sensor. Furthermore, they can even compensate for its limitation such as restricted field-of-view (which makes a part of the surface of the visuotactile sensor unusable) and measured depth dropout errors due to over-saturation in close ranges. 

Finally, in our \textbf{fourth contribution}, we demonstrate the utility of the proposed model in a depth-based tactile deformation and pose estimation application. The performance when using predicted depth images compares favourably against the current depth-sensor-captured information. We then discuss the implications of our work in future soft tactile sensor designs - for instance in permitting greater flexibility in manipulator design and in facilitating more wide-spread tactile sensor placement on soft-robot structures.



\section{Related Work}
\subsection{Visuotactile Sensing of Depth}
Visuotactile sensors that can capture high resolution tactile depth maps include examples such as \emph{GelSight}~\cite{gelsight,gelsight2}, \emph{GelSlim}~\cite{bauza2019tactile}, and \emph{Soft-bubbles}~\cite{Alspach2019}. As noted in Sec. \ref{sec:intro}, three broad classes of methods have been deployed towards capturing tactile depth maps - direct capture with internal depth imaging sensors (e.g., \emph{Soft-bubbles}~\cite{Alspach2019}), model based methods that rely photometric stereo (e.g., GelSight~\cite{dong2017}), and by using machine learning to infer depth from visuotactile RGB images (e.g., GelSlim~\cite{bauza2019tactile}).

These methods, however, require careful calibration, specific conditions and are usually constrained to models trained for single objects. GelSight~\cite{gelsight,gelsight2} requires specific lighting from multiple LEDs to generate depth maps using a photometric stereo algorithm. GelSlim~\cite{bauza2019tactile} requires a specific training protocol involving multiple objects placed at different positions of the tactile sensor in order to generate local shape estimates and the corresponding height maps. Furthermore, methods that directly reconstruct 3D shapes from images, such as \cite{pointset,genre}, require segmentation masks to isolate the object before reconstruction takes place. Our approach densely reconstructs the entire image on a per-pixel basis, can be trained with input-output image and depth maps without prior shape knowledge or specific protocol, and as shown in experiments generalizes well to unseen objects.  



\subsection{Monocular Depth Estimation}


Estimating depth from a single image is inherently an ill-posed problem, since there are infinite possible world states that could have generated it. However, the seminal work of Eigen et al.~\cite{eigen2014depth} has shown that it is possible to train a neural network to learn appearance-based features capable of outputting a dense depth map, containing per-pixel distance estimates. Following~\cite{eigen2014depth}, a substantial amount of work has been done to improve the accuracy and performance of supervised monocular depth estimation, including the use of Conditional Random Fields (CRFs)~\cite{depthcrf}, different loss functions~\cite{huberloss,packnet-semisup}, joint multi-task optimization~\cite{normalscvpr2,selfsupsem,packnet-semguided}, representation in different domains~\cite{fouriercvpr}, formulating depth estimation as an ordinal classification problem~\cite{dorncvpr} and the use of local planar guidance layers during the upsampling stage~\cite{lee2019big}. At the same time, self-supervision has emerged as a way to train models without ground-truth information at training time. Instead of using ground truth depth information, geometric priors are used to constrain learning in such a way that depth emerges as a proxy task for the projection of information between two images, either spatially (i.e. from stereo cameras)~\cite{pillai2018superdepth, zhou2018stereo, ummenhofer2017demon} or temporally (i.e. from a monocular at different time-steps)~\cite{packnet,godard2018digging2,vijayanarasimhan2017sfm}. 

However, the vast majority of these works consider texture-rich scenes with large depth ranges (Figure \ref{fig:teaser}), usually outdoor driving scenarios with ground-truth collected using LiDAR \cite{geiger2013vision} or indoor environments with input-output pairs collected using RGBD sensors \cite{sunrgbd}. Little attention has been given to monocular depth estimation from very small depth ranges, where the input images contain little to no texture information. In this work we explore how various depth networks commonly used in these texture-rich and large depth range scenarios perform under these more challenging conditions, including generalization experiments between different objects and cameras. 

\section{Methodology}

\begin{figure*}[t!]
\vspace{5mm}
\includegraphics[width=\linewidth]{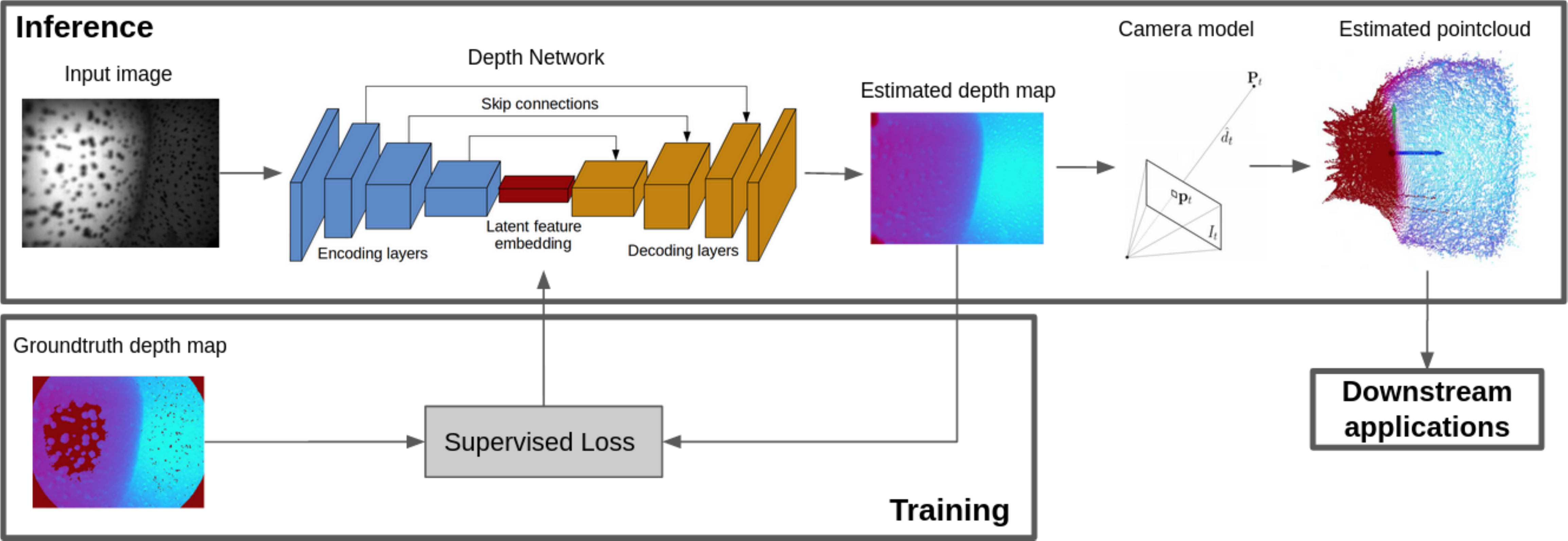}
\caption{Diagram of our proposed framework. We use as input a $H \times W$ grayscale image, that is processed by our \textit{Depth Network} to produce an estimated depth map of the same resolution. At training time, the corresponding ground-truth depth map is also provided, and the \textit{Depth Network} is optimized as to minimize the difference between estimated and ground-truth depth maps, given an objective function (Equation~\ref{eq:loss}). The estimated depth map is lifted to 3D using the camera model, to produce an estimated pointcloud that can then be used for downstream applications (e.g. pose estimation).}
\label{fig:framework}
\vspace{-3mm}
\end{figure*}

\subsection{Monocular Depth Estimation}

The aim of monocular depth estimation is to learn a function $f_D:I \to D$ that recovers the depth $\hat{D}=f_D\left(I(p)\right)$ for every pixel $p\in I$.  While a number of works have been proposed that learn to regress $f$ in the self-supervised regime using only videos as supervision~\cite{zhou2018stereo,godard2018digging2}, because we have access to ground truth depth information acquired by the gripper's IR/depth camera we can treat the problem of learning to regress depth from images purely as a supervised learning problem. We define depth estimator $f_D$ parameterized by $\theta_D$ as:
\begin{equation} \label{eq:depth_model}
\begin{gathered}
\hat{\theta}_D = \argmin_{\theta_D}\mathcal{L}\left(I,D;\theta_D\right)
\end{gathered},
\end{equation}
where the objective function to be minimized is given by:
\begin{equation} \label{eq:loss}
\begin{gathered}
\mathcal{L}_D(I,D)=\mathcal{L}_{supp}\left(I,D\right)\odot M_{supp}\left(D\right).
\end{gathered}
\end{equation}

\textbf{Supervised loss.}~~We impose a per-pixel loss between the predicted $\hat{D}$ and ground-truth $D$ depth maps as a way to optimize the neural network weights at training time. Although different supervised losses have been proposed \cite{huberloss, dorncvpr}, in our experimental evaluation we focus on a simple L1 function $\mathcal{L}_{supp} = \|\hat{D} - D\|_1$, and leave the exploration of how performance can be further improved by leveraging different losses to future work . The term $M_{supp}$ denotes a per-pixel mask and is used to remove pixels with invalid depth from the optimization, with $\odot$ denoting element-wise matrix multiplication. This mask enables training using sparse supervision while still producing dense pixel-wise depth predictions.

\subsubsection{Depth Networks}

We experiment with three different network architectures, each having been demonstrated as effective under different circumstances: ResNet~\cite{he2016deep}, BTS~\cite{lee2019big} and PackNet~\cite{packnet}. They are all encoder-decoder architectures, with skip connections to facilitate gradient propagation, and estimate dense inverse depth maps from input images. For more details on the architecture of these networks please refer to~\cite{he2016deep,packnet,lee2019big}. 

\textbf{ResNet.} As shown in~\cite{godard2018digging2}, with ImageNet~\cite{deng2009imagenet} pretraining, this light-weight network is well suited for monocular depth prediction.

\textbf{PackNet.} This architecture leverages specially designed packing and unpacking blocks and achieves state-of-the-art depth estimation results~\cite{packnet}. It was crafted to exploit small texture variations, leading to high-quality depth predictions.


\textbf{BTS.} As shown in~\cite{lee2019big}, this architecture has achieved state-of-the-art results in the KITTI outdoor dataset~\cite{geiger2013vision}. It uses local planar guidance layers, thus ensuring that upsampled features share the same predicted surface normal vectors.


Each network receives as input a $224 \times 224$ grayscale image, and outputs a $224 \times 224$ image with per-pixel metric depth estimates.  We also include inference times for each network (batch size of $1$, measured on a Titan V100 GPU card), with \textit{ResNet} achieving the fastest performance, with 58 fps (frames per second); followed by \textit{PackNet}, with 24 fps; and finally \textit{BTS}, with 20 fps. Figure~\ref{fig:framework} shows a high level diagram of the proposed learning framework.

\subsection{Visuotactile Sensor Configuration and Test Objects}
\label{subsubsec:visuotactile}

\begin{figure}[t!]
\centering
\includegraphics[width=0.72\columnwidth,height=5.25cm]{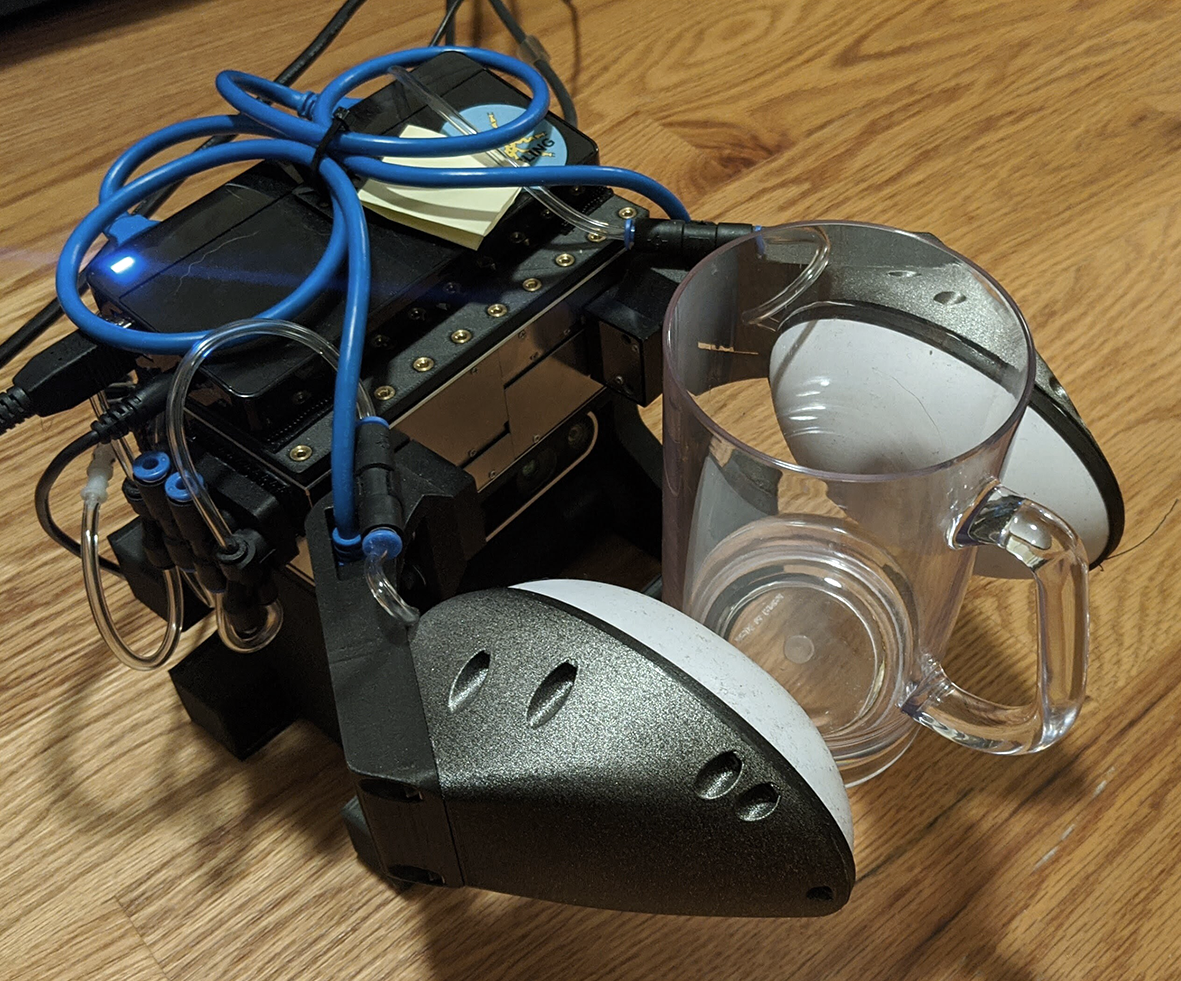}
\caption{Ellipsoidal, \emph{Soft-bubble} gripper used to gather data, contacting a transparent mug.}
\label{fig:gripper_bubble}
\end{figure}

For the purposes of gathering test data, we used a gripper with 2 ellipsoidal form-factor \emph{Soft-bubble} sensor fingers, as shown in Figure \ref{fig:gripper_bubble}. Each of the gripper fingers is independently inflated and houses its own camera. There is no internal light source in these sensors barring the IR emitter in the ToF camera. Additionally, each \emph{Soft-bubble} has a black pseudo-random dot pattern printed on the interior of the compliant membrane for adding visual texture on the surface~\cite{kuppuswamy_soft-bubble_2020}. 


\begin{figure}[t!]

\centering
\subfloat{
\includegraphics[width=0.20\columnwidth,keepaspectratio]{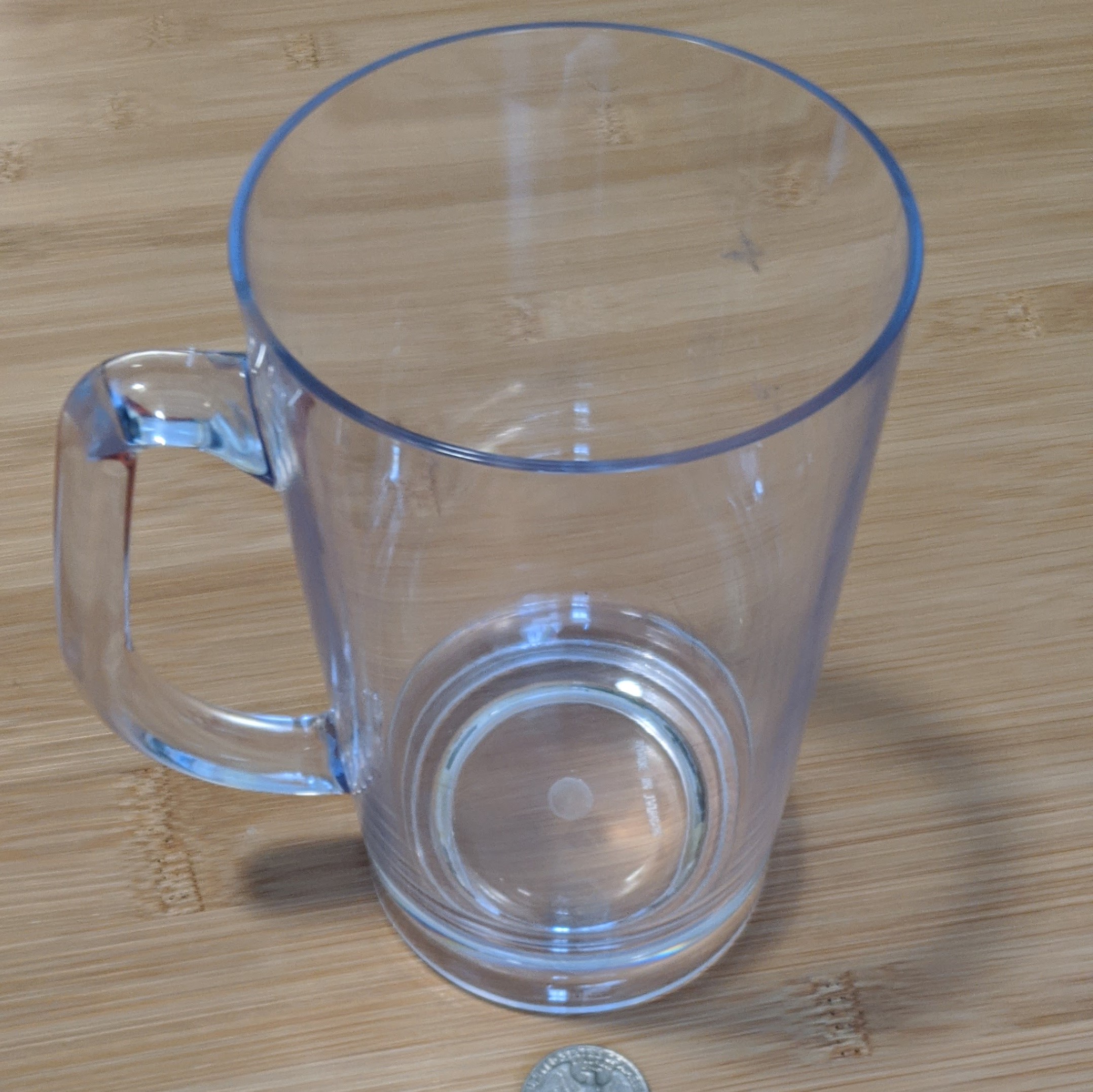}
}
\subfloat{
\includegraphics[width=0.20\columnwidth,keepaspectratio]{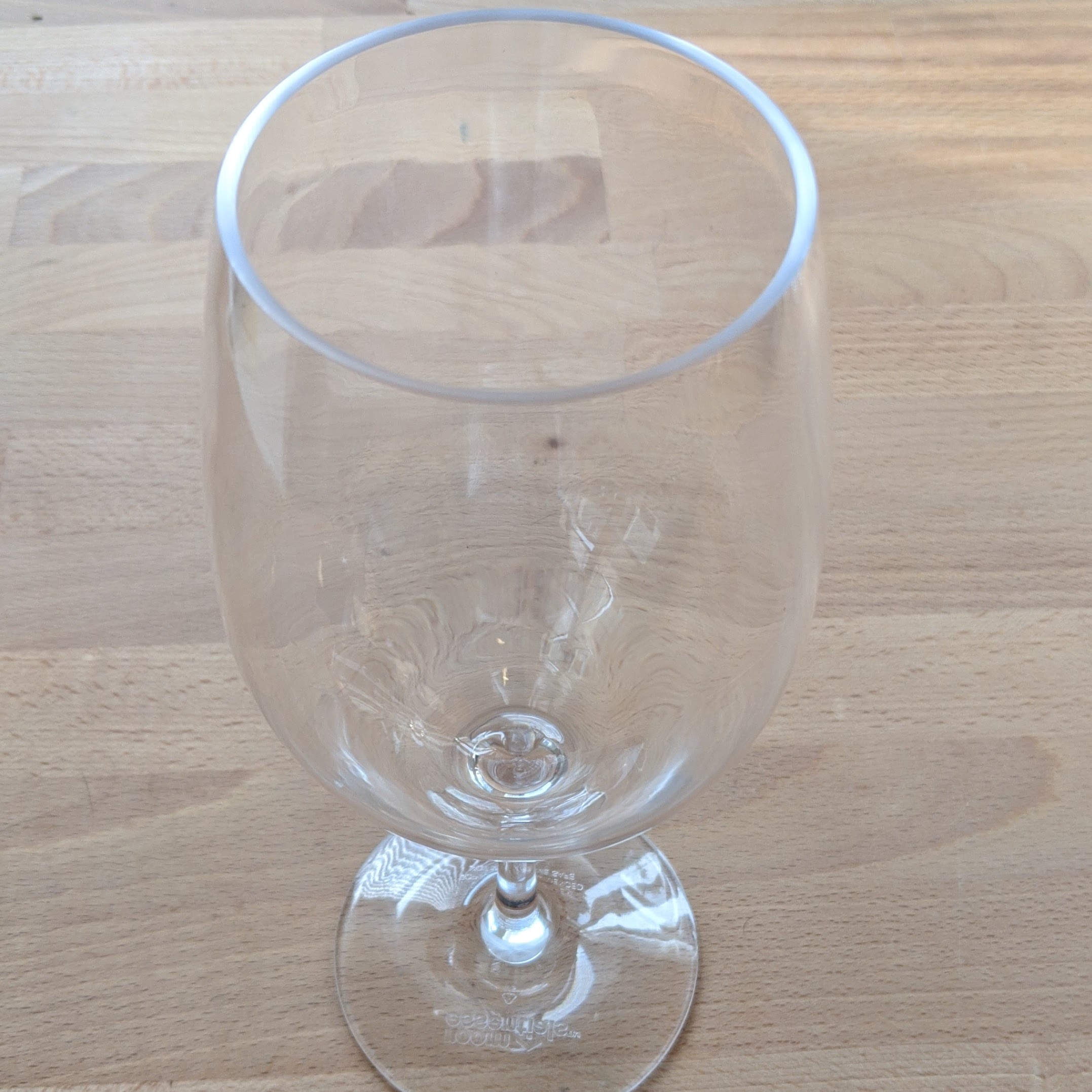}
}
\subfloat{
\includegraphics[width=0.20\columnwidth,keepaspectratio]{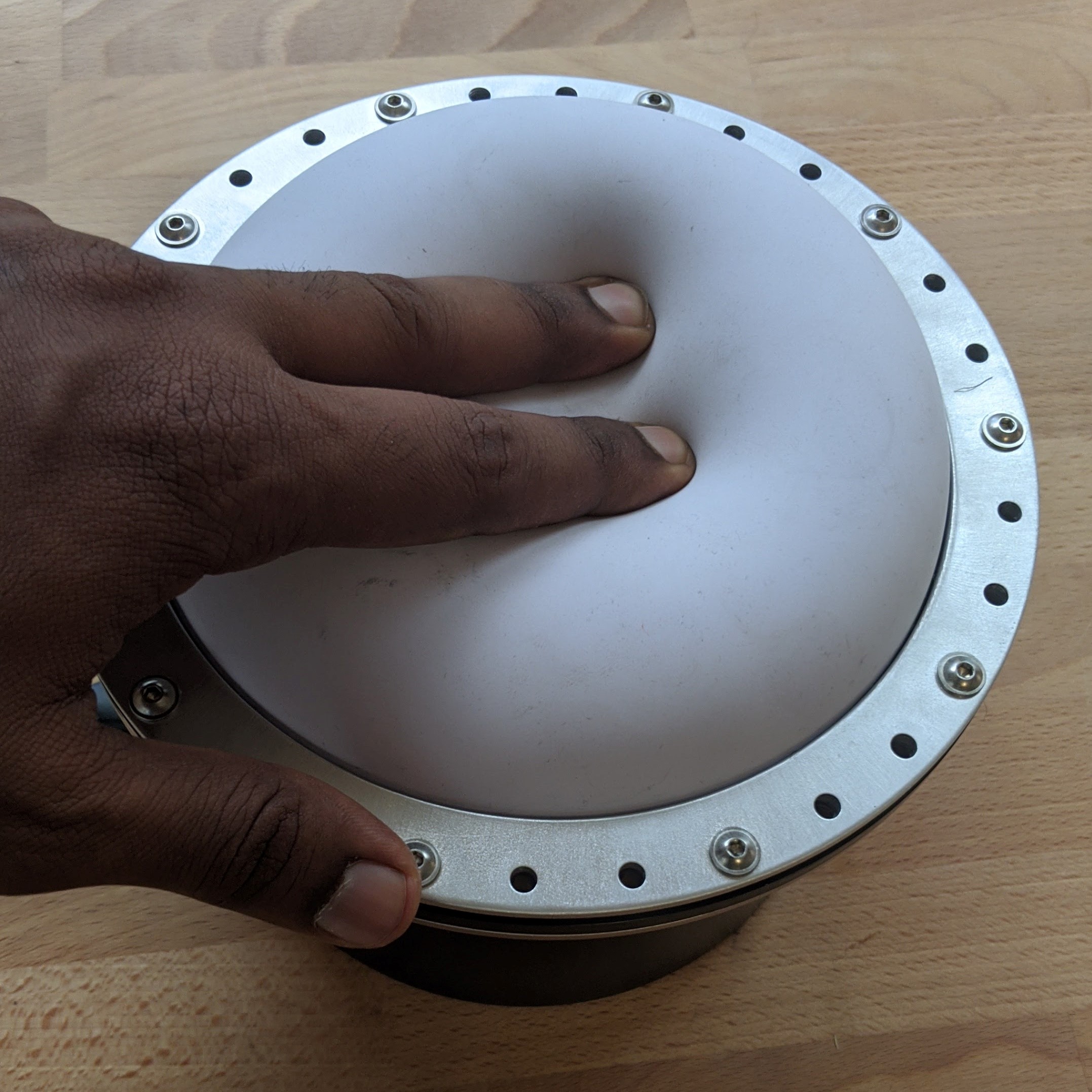}
}
\subfloat{
\includegraphics[width=0.20\columnwidth,keepaspectratio]{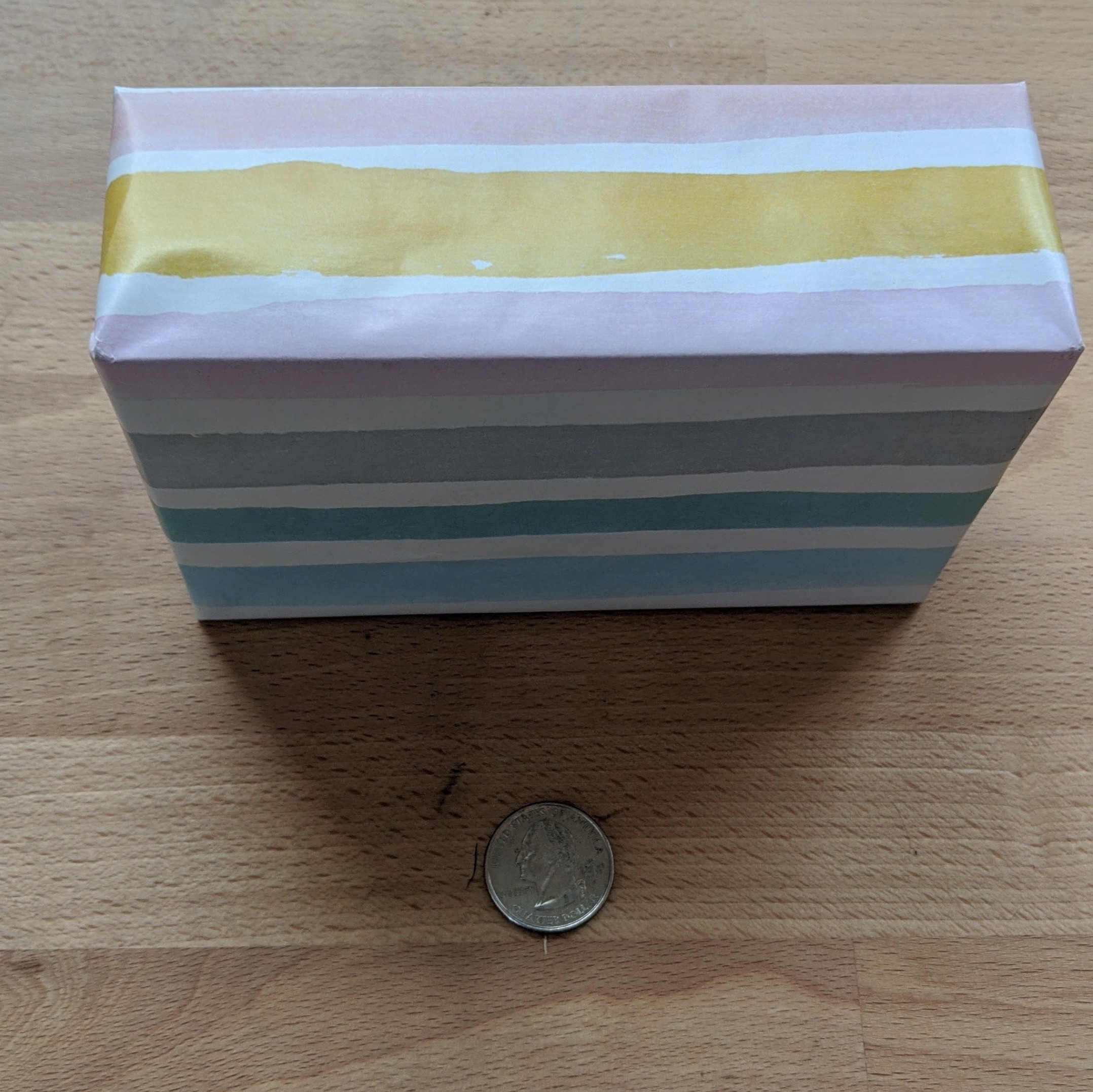}
}
\setcounter{subfigure}{0}
\subfloat[Mug]{
\includegraphics[width=0.20\columnwidth,keepaspectratio]{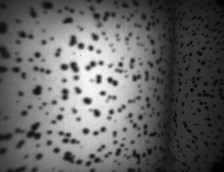}
\label{fig:object_mug}
}
\subfloat[Wine Glass]{
\includegraphics[width=0.20\columnwidth,keepaspectratio]{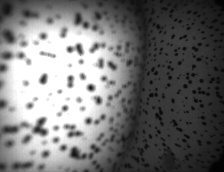}
\label{fig:object_wine_glass}
}
\subfloat[Fingers]{
\includegraphics[width=0.20\columnwidth,keepaspectratio]{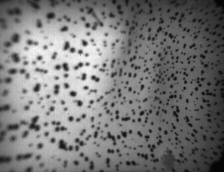}
\label{fig:object_fingers}
}
\subfloat[Box]{
\includegraphics[width=0.20\columnwidth,keepaspectratio]{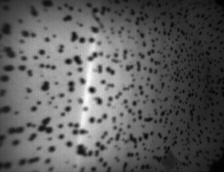}
\label{fig:object_box}
}

\caption{Test Objects with respective images from the IR camera within the Gripper Sensor}
\label{fig:objects}

\end{figure}

To generate data with these visuotactile sensors, one of four objects is pressed against the surfaces to simulate a touch or grasp of the object as might be representative of sensing seen on a robotic end effector or gripper. The test objects used were a glass mug (Figure \ref{fig:object_mug}), a wine glass (\ref{fig:object_wine_glass}), fingers (\ref{fig:object_fingers}), and a small rectangular box (\ref{fig:object_box}). Each of these objects is shown in Figure \ref{fig:objects} above a respective example IR camera image obtained from within the compliant sensor when the object was in contact. The objects clearly deform the printed dot pattern and affect how the membrane reflects the internal light and also vary the depth returns.

\section{Experimental Results}

\subsection{Datasets}
\label{subsec:datasets}


To validate our method we perform experiments with the \emph{Soft-bubble} gripper that contains contains a printed internal dot pattern (Figure \ref{fig:gripper_bubble}). We collect five different datasets: three while manipulating household objects with different geometries (a wine glass, a mug, and a box); one while manually poking the \emph{Soft-bubble} surface; and finally when no objects are manipulated. We note that data is collected from both the left and right cameras housed inside the gripper (as described in Section~\ref{subsubsec:visuotactile}). The total number of images collected is $21960$, with details in Table~\ref{table:dotted_datasets}. To evaluate \textit{in-class} performance, we create a random $70/30$ partition of each dataset. The resulting training split contains $15372$ images, while the test split contains $6588$ images (see Section~\ref{subsubsec:in_class}). To evaluate \textit{out-of-class} generalization, we perform leave-one-out experiments, i.e. train on three of the four datasets containing objects and evaluate on the remaining one. For example, when evaluating on the Wine Glass dataset, the training split contains $16140$ images while the test split contains $5820$ (see Section~\ref{subsubsec:cross_object}). Finally, we also evaluate generalization across cameras, by using one as the training split and another as the test split. For this experiment both the train and the test splits consist of $10980$ images (see Section~\ref{subsubsec:cross_cam}).

\begin{table}[!b]
\centering
\begin{tabular}{l|c|c|c|c|c}
& Wine Glass & Box & Fingers & Mug & No contact \\
\toprule
\# images & 5820 & 6038 & 4752 & 1600 & 3750 \\
\bottomrule
\end{tabular}
\vspace{2mm}
\caption{Data collected using the \emph{Soft-bubble} gripper configuration.}
\label{table:dotted_datasets}
\end{table}

\subsection{Training}
\label{subsec:training}

Our models were implemented using PyTorch~\cite{paszke2017automatic} and trained across 8 Titan V100 GPUs. We use the Adam optimizer~\cite{kingma2014adam}, with $\beta_1=0.9$ and $\beta_2=0.999$. The depth networks were reproduced based on their official implementations, and trained for a total of $100$ epochs, with a batch size of 4 samples per GPU and learning rate of $2 \cdot 10^{-4}$, which is halved after every 40 epochs. 




\subsection{Monocular Depth Estimation}
\label{subsec:results_depth}
\subsubsection{Metrics}

To evaluate the performance of our depth networks we used the standard metrics found in the literature, which include 
\textbf{Abs. Rel}: $\frac{1}{N} \sum_{\tilde{d} \in \tilde{D}} \frac{|\tilde{d} - d|}{\tilde{d}}$, 
\textbf{RMSE}: $\sqrt{\frac{1}{N} \sum_{\tilde{d} \in \tilde{D}} |\tilde{d} - d|}$, 
\textbf{RMSElog}: $\sqrt{\frac{1}{N} \sum_{\tilde{d} \in \tilde{D}} |\log \tilde{d} - \log d|}$, 
\textbf{SILog}: $\frac{1}{N} \sum_{\tilde{d} \in \tilde{D}} \left( \log d - \log \tilde{d} \right)^2 - \frac{1}{N^2} \left(\sum_{\tilde{d} \in T} \log d - \log \tilde{d} \right)^2$ and \textbf{Accuracy} $\delta(thr) = \text{\% of s.t. } \max\left(\frac{\tilde{d}}{d}, \frac{d}{\tilde{d}}\right) < thr$. In these, $\tilde{d}$ and $d$ represents respectively ground-truth and corresponding predicted depth values, with $\tilde{D}$ being the set containing all valid ground-truth depth pixels. Note that, for Accuracy, we use powers of $1.05$ as thresholds, instead of $1.25$ as is standard in the literature. We found that this smaller threshold is more representative due to the smaller depth ranges that are considered in this work.
 

\subsubsection{In-Class Generalization}
\label{subsubsec:in_class}

Table~\ref{table:depth-in-class} summarizes our results when training and testing on the same set of objects, as described in Section~\ref{subsec:datasets}. We perform experiments with the \textit{ResNet}, \textit{PackNet} and \textit{BTS} architectures and note that when trained and evaluated with the same object classes all networks achieve similar performance, with ResNet performing slightly better. With an RMSE error of $0.8cm$ we conclude that the networks successfully map the tactile skin deformations observed in the IR image to per pixel depth.

\begin{figure}[t!]
\vspace{-5mm}
\centering
\subfloat[Num valid depths]{
\includegraphics[width=0.3\columnwidth,keepaspectratio]{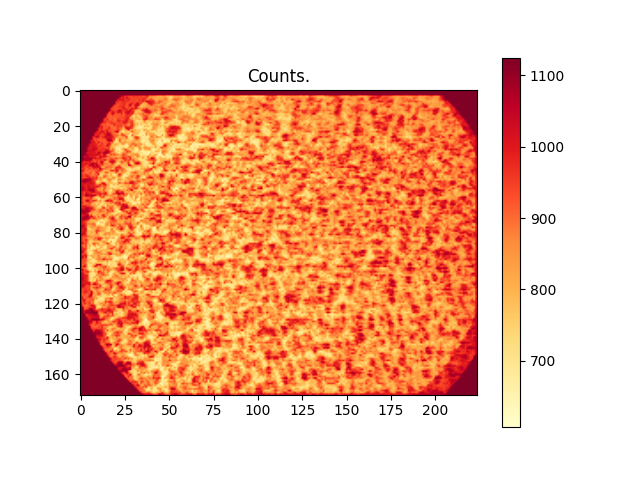}
\label{fig:dotless_input}
}
\subfloat[Error mean]{
\includegraphics[width=0.3\columnwidth,keepaspectratio]{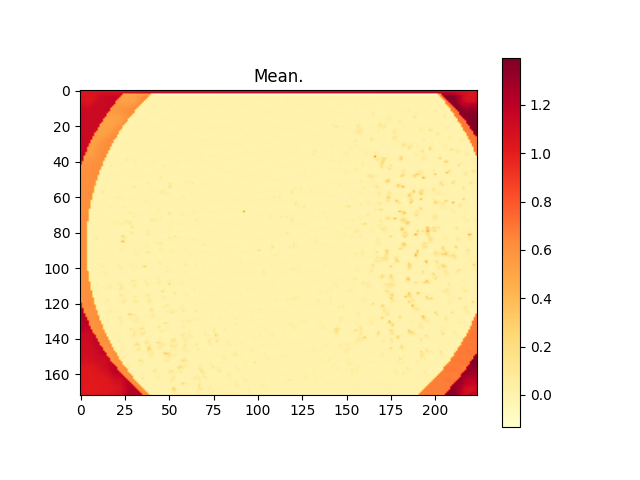}
\label{fig:dotless_gt}
}
\subfloat[Error variance]{
\includegraphics[width=0.3\columnwidth,keepaspectratio]{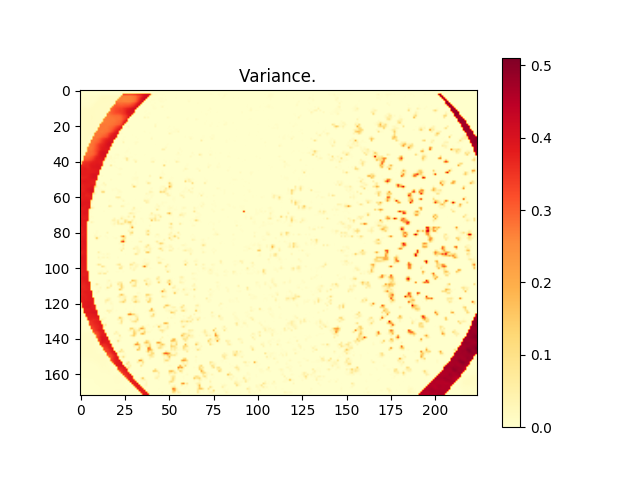}
\label{fig:dotless_pred}
}
\caption{We compute the count of valid depth values per-pixel, as well as the error mean and variance between model predictions and ground truth over the "No contact" dataset.}
\vspace{-2mm}
\label{fig:stats}
\end{figure}

Additionally, We perform an in-depth statistical analysis of the performance of the \textit{PackNet} model. We compute the per-pixel mean and variance of the predicted depth and the ground truth depth over the \textit{No contact} dataset test split. Fig~\ref{fig:stats} shows these results, as well as the number of valid depth measurements per-pixel. We note that the mean and variance of errors is low, with the mean being high on the corners of the images, where the imaging sensor returns invalid measurements; this might exist due to material changes between the latex membrane to the housing shear resulting in ground truth sensor dropouts. In addition, we note a number of "spots" with higher errors, which are correlated with the position of the dot pattern on the tactile skin.

\subsubsection{Out-of-Class Generalization}
\label{subsubsec:cross_object}

We measure our depth network's ability to generalize across different objects by employing a \textit{leave-one-out} strategy, i.e. training on a subset of available datasets and evaluating on the remaining one. Since each dataset is composed of observations taken from a single object, this protocol enables us to measure the degradation in depth estimates when inference is performed on objects that were not observed at training time. The results for the different depth networks are shown in Table \ref{tab:cross-object}.  We can see that the \textit{PackNet} architecture consistently outperforms \textit{ResNet} and \textit{BTS} in all considered metrics. This is aligned with the observation that \textit{PackNet} is designed to preserve fine-grained visual information, which is especially beneficial in this low-texture scenario, and also it has shown better generalization properties \cite{packnet}. Interestingly, this behavior was not observed in the \textit{in-class} generalization experiments, which indicates that there are significant variations in the visual patterns depending on which object is being pressed against the \emph{Soft-bubble} gripper.


\begin{table}[t!]
\renewcommand{\arraystretch}{0.87}
\centering
{
\small
\setlength{\tabcolsep}{0.3em}
\begin{tabular}{l|ccccc}
\toprule
\multicolumn{6}{c}{\textbf{In-Class Generalization}} \\
\toprule
\textbf{Method} & 
Abs.Rel $\downarrow$ &
RMSE$\downarrow$ &
RMSE$_{log}$$\downarrow$ &
SILog$\downarrow$ & 
$\delta_{1.05}$$\uparrow$ \\
\toprule

PackNet~\cite{packnet} & 0.018 & 0.008 & 0.026 & 2.595 & 0.947  \\

BTS~\cite{lee2019big} & 0.018   & 0.008   &  0.025   & 2.457 & 0.948 \\

ResNet~\cite{he2016deep} & 0.017 & 0.008 & 0.023 & 2.336 & 0.955 \\

\bottomrule

\end{tabular}
}
\caption{Monocular depth estimation results for in-class generalization, using different networks.}
\label{table:depth-in-class}
\end{table}









\begin{table}[t!]
\renewcommand{\arraystretch}{0.87}
\centering
{
\small
\setlength{\tabcolsep}{0.3em}
\begin{tabular}{c|c|ccccc}
\toprule
\multicolumn{7}{c}{\textbf{Out-of-Class Generalization}} \\
\toprule
& \textbf{Dataset} & 
Abs.Rel &
RMSE &
RMSE$_{log}$ &
SILog & 
$\delta<1.05$
\vspace{0.5mm}\\
\toprule

\parbox[t]{2mm}{\multirow{4}{*}{\rotatebox[origin=c]{90}{\textit{ResNet}}}}
& \textit{Mug} & 0.036 & 0.013 & 0.046 & 4.317 & 0.735 \\
& \textit{Wine Glass} & 0.040 & 0.015 & 0.051 & 4.889 & 0.692 \\
& \textit{Fingers} & 0.040 & 0.018 & 0.056 & 5.395 & 0.715 \\
& \textit{Box} & 0.051 & 0.019 & 0.063 & 6.102 & 0.618 \\
\midrule
\parbox[t]{2mm}{\multirow{4}{*}{\rotatebox[origin=c]{90}{\textit{BTS}}}}
& \textit{Mug} & 0.038 & 0.015 & 0.050 & 4.629 & 0.744 \\
& \textit{Wine Glass} & 0.038 & 0.015 & 0.050 & 4.579 & 0.730 \\
& \textit{Fingers} & 0.039 & 0.018 & 0.056 & 5.504 & 0.741 \\
& \textit{Box} & 0.048 & 0.018 & 0.062 & 5.899 & 0.673 \\
\midrule
\parbox[t]{2mm}{\multirow{4}{*}{\rotatebox[origin=c]{90}{\textit{PackNet}}}}
& \textit{Mug} & 0.034 & 0.013 & 0.042 & 3.458 & 0.765 \\
& \textit{Wine Glass} & 0.034 & 0.014 & 0.047 & 4.338 & 0.769 \\
& \textit{Fingers} & 0.030 & 0.014 & 0.047 & 4.479 & 0.832 \\
& \textit{Box} & 0.047 & 0.019 & 0.061 & 5.687 & 0.652 \\
\bottomrule


\end{tabular}
}
\caption{Monocular depth estimation results for cross-dataset generalization using different networks. Independent models trained on 3/4 object datasets, evaluated on the remaining one.}
\label{tab:cross-object}
\end{table}


\begin{table}[t!]
\renewcommand{\arraystretch}{0.87}
\centering
{
\small
\setlength{\tabcolsep}{0.3em}
\begin{tabular}{c|c|ccccc}
\toprule
\multicolumn{7}{c}{\textbf{Cross-camera generalization (PackNet)}} \\
\toprule
& \textbf{Dataset} & 
Abs.Rel$\downarrow$ &
RMSE$\downarrow$ &
RMSE$_{log}$$\downarrow$ &
SILog$\downarrow$ & 
$\delta_{1.05}$$\uparrow$ \vspace{0.5mm}\\
\toprule

\parbox[t]{2mm}{\multirow{4}{*}{\rotatebox[origin=c]{90}{\textit{Left}}}}
& \textit{Mug}
& 0.032 & 0.015 & 0.051 & 4.944 & 0.820 \\
& \textit{Wine Glass}
& 0.034 & 0.015 & 0.053 & 5.177 & 0.804 \\
& \textit{Fingers}
& 0.047 & 0.020 & 0.068 & 6.392 & 0.628 \\
& \textit{Box}
& 0.052 & 0.020 & 0.072 & 6.442 & 0.606 \\

\midrule
\parbox[t]{2mm}{\multirow{4}{*}{\rotatebox[origin=c]{90}{\textit{Right}}}}

& \textit{Mug}
& 0.031 & 0.012 & 0.041 & 3.871 & 0.803 \\
& \textit{Wine Glass}
& 0.046 & 0.020 & 0.058 & 5.230 & 0.595 \\
& \textit{Fingers}
& 0.062 & 0.026 & 0.077 & 4.484 & 0.389 \\
& \textit{Box}
& 0.058 & 0.023 & 0.071 & 6.180 & 0.459 \\

\bottomrule

\end{tabular}
}
\caption{Dataset transfer results using PackNet \cite{packnet}. Independent models were trained using input-output pairs from each of the available cameras, and evaluated on the remaining one.}
\label{tab:cross-camera}
\end{table}

\subsubsection{Cross-Camera Generalization}
\label{subsubsec:cross_cam}

Similarly, we also performed experiments to examine our depth network's ability to generalize between different cameras, by training using only information from one single camera and evaluating on the remaining one. Results are shown in Table \ref{tab:cross-camera} for the \textit{PackNet} architecture, which has shown better performance in the \textit{out-of-class} experiments described in the previous section. From these results we can see that camera generalization poses a bigger challenge than object generalization. We hypothesize that this is because of the texture-less nature of the observed surface, that makes geometric patterns and lighting variations a key part of the appearance-based features learned by the network. Changing the camera viewpoint also changes these observed geometric and lighting patterns, which in turn degrades the performance of the depth network.

\subsubsection{Qualitative results}

\begin{figure}[t!]
\centering
\subfloat{
\includegraphics[width=0.11\textwidth,height=1.8cm]{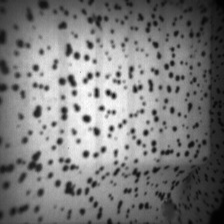}
} 
\subfloat{
\includegraphics[width=0.11\textwidth,height=1.8cm]{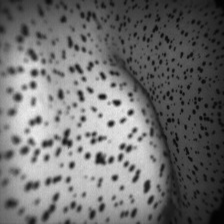}
} 
\subfloat{
\includegraphics[width=0.11\textwidth,height=1.8cm]{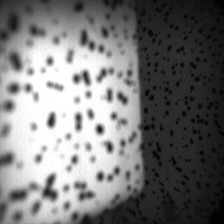}
} 
\subfloat{
\includegraphics[width=0.11\textwidth,height=1.8cm]{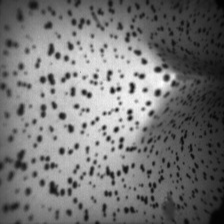}
} 
\\ \vspace{-3mm}
\subfloat{
\includegraphics[width=0.11\textwidth,height=1.8cm]{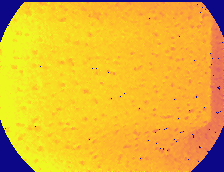}
} 
\subfloat{
\includegraphics[width=0.11\textwidth,height=1.8cm]{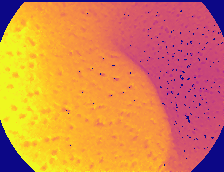}
} 
\subfloat{
\includegraphics[width=0.11\textwidth,height=1.8cm]{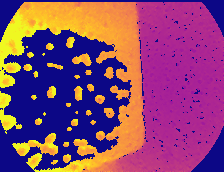}
} 
\subfloat{
\includegraphics[width=0.11\textwidth,height=1.8cm]{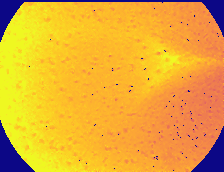}
} 
\\ \vspace{-3mm}
\subfloat{
\includegraphics[width=0.11\textwidth,height=1.8cm]{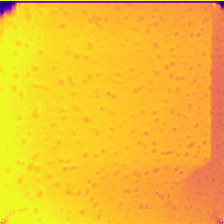}
} 
\subfloat{
\includegraphics[width=0.11\textwidth,height=1.8cm]{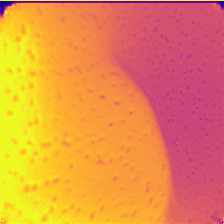}
} 
\subfloat{
\includegraphics[width=0.11\textwidth,height=1.8cm]{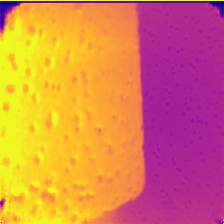}
} 
\subfloat{
\includegraphics[width=0.11\textwidth,height=1.8cm]{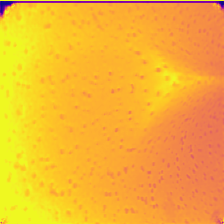}
} 
\\
\caption{Qualitative depth estimates (bottom row) obtained using \textit{PackNet} on the \emph{Soft-bubble} gripper data (test images in top row, not seen during training).  Note that even though the ground-truth depth maps (middle row) may be sparse and noisy, the predicted depth maps are still dense and smoothly complete missing areas, while maintaining sharpness in regions with sudden depth discontinuities.}
\label{fig:qualitative}
\vspace{-3mm}
\end{figure}

Examples of predicted depth maps produced by \textit{PackNet} on held out test input IR images are shown in Figure \ref{fig:qualitative} together with the ground-truth depth. These qualitative results indicate that the predicted depth maps closely approximate the ground-truth depth information, including finer details such as small holes in the \emph{Soft-bubble} gripper and sudden depth discontinuities where deformation is particularly large. At the same time, the predicted depth maps are capable of smoothing out areas where there is no ground-truth information to produce dense, per-pixel estimates regardless of depth ranges.

\section{Proximity Pose Estimation}
As a demonstration of a downstream application using the estimated depth, we utilized an optimization-based pose estimation pipeline similar to Kuppuswamy et al \cite{kuppuswamy_soft-bubble_2020} which in turn is broadly similar to other optimization based pose estimation methods such as in Izatt et al \cite{Izatt2017}. The first stage of this pipeline is a depth-based contact patch estimator in which we use the depth from an uncontacted tactile sensor to compute a contact patch mask that selects only the parts of the point cloud that are in direct contact with the object. The output of this masked point-cloud is then fed into a nonlinear optimization that estimates the pose ${^G}X_T$ (the pose of the chosen geometry $G$ in the tactile sensor frame $T$) as:

\begin{equation}
\begin{split}
\theta^* = &\argmin_{\theta} \sum_{i=0}^{N} \left[ \phi({^G}X_T(\theta)\; ^T{p_i}{^Q}) \right]^2, \quad  \text{s.t.}: \theta \in \mathrm{SE}(3),
\end{split}
\label{eq:proximity_pose_optimization}
\end{equation}
where $\theta$ is the chosen parameterization of the pose ${^G}X_T$, for a $7$-dimensional vector composed of the three translational coordinates and four rotational coordinates (represented by a quaternion). We use a unit-norm constraint on the quaternion part of $\theta$. The function $\phi(.)$ is a proximity field, which is a generalization of the notion of a signed distance field. The masked point cloud, is represented in Eq. \ref{eq:proximity_pose_optimization} by $ ^T{p_i}{^Q}$.

\textbf{Experiments and Results:}
Similar to previous work \cite{kuppuswamy_soft-bubble_2020}, we test on cylindrical mugs by computing pose-estimates for cylinders. Note that the mug that was tested was not part of the set of objects used for training the depth estimation network. The masked pointclouds were downsampled to $1500$ points and the pose estimates were computed at the rate of $5-10$ Hz. The pose estimator is susceptible to errors due to local-minima; a sufficiently close initial condition was generated \textit{apriori} corresponding to an upright mug in the world with respect to the gripper and tactile sensors.

\begin{figure}[t!]
\begin{tabular}{cc}
\raisebox{20mm}{\multirow{2}{*}{
\subfloat[Pose estimation on monocular depth maps]{
\includegraphics[width=0.50\columnwidth]{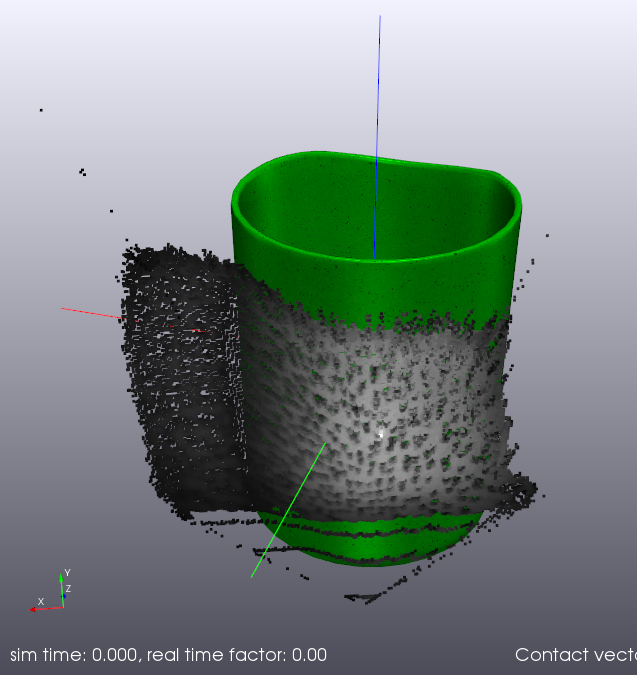}}
}
}
& 
\hspace{-5mm} \subfloat[Position norm error histogram]{
\includegraphics[width=.40\columnwidth]{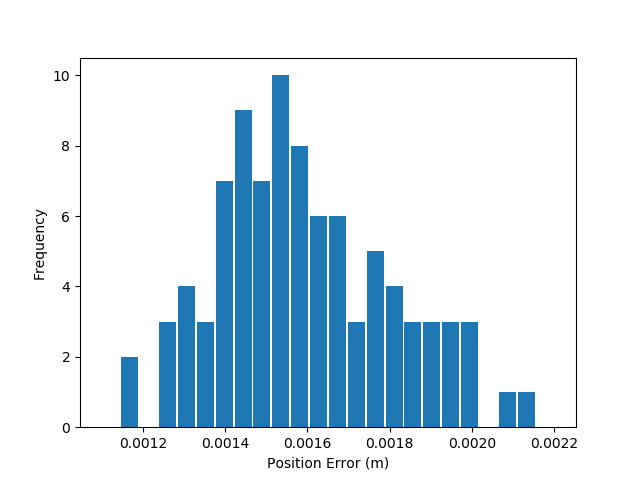}
}
\\
&
\hspace{-5mm} \subfloat[Orientation error histogram]{
\includegraphics[width=.40\columnwidth]{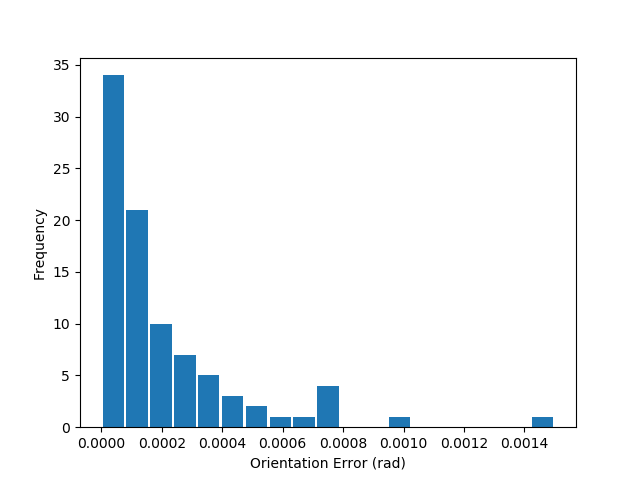}
}
\end{tabular}

\caption{Pose estimation experimental results. (a) In-hand pose estimation of a cylinder using estimated depth images (b) Histogram of position norm difference of computed pose on estimated point-clouds vs. ground-truth point-clouds, and (c) Histogram of orientation error of computed pose on estimated point-clouds vs. ground-truth point-clouds}
\label{fig:pose_estimation}
\end{figure}

To compute the pose estimates, we selected a subset of 91 images from the mug dataset where the object is held by the gripper equipped with the \emph{Soft-bubble} visuotactile sensors. The mug pose was perturbed by hand. For the experiments reported here, we used images that were obtained with little to no movement of the object to enable us to compare the estimated depth versus the raw ground-truth depth without bias introduced due to initialization of the pose estimator. The pose estimation errors between estimated and ground-truth depth images are shown in Fig. \ref{fig:pose_estimation}. 


The position error between the pose estimate on the estimated depth ($[p_1^T, q_1^T]^T$) and on the ground-truth ($[p_2^T, q_2^T]^T$) was computed as a norm position difference $e_{x} = \| p_1 - p_2 \|$, whereas the orientation error was computed by $e_{\theta} = 1 - |q_1.q_2|$ as defined by Huynh \cite{huynh2009metrics} (Eq. 20). The histogram of computing this difference for the $91$ test saples is shown in Fig. \ref{fig:pose_estimation}. The median/mean/variance of the \textbf{pose estimation errors} between the sensor generated (ground-truth) and the predicted depth are $[0.16, 0.19, 1.14e-04]$ for the object position (in cm) and $[0.0005, 0.0010, 2.75e-06]$ for its orientation (in radians). 
The errors are well within the tolerances of in-hand pose estimation for end-to-end manipulation.


\section{Conclusion}
In this paper, we show that deep networks for monocular depth estimation can be trained to operate at three orders of magnitude shorter ranges than existing use cases, from meters to millimeters. Our real-world experiments with \emph{Soft-bubble} visuotactile sensors show that we can robustly estimate deformation geometry and contact patches from a single internal IR camera and a deep convolutional network. We propose a sample-efficient learning system trained from only a few object categories and contact configurations. Our model is accurate and generalizes across different object types and sensor configurations.
Our results open up many possibilities for a variety of close-proximity sensors (tactile sensing skins and beyond, e.g., wrist-mounted cameras) with vastly simplified and compact imaging systems, rendering low-cost, mechanically robust, high-resolution tactile sensing for soft robotics viable at scale.


{\small
\bibliographystyle{ieee}
\bibliography{references}
}
\end{document}